%% file: main.tex
\documentclass[letterpaper]{article}
\usepackage{aaai2026} 
\usepackage{times} 
\usepackage{helvet} 
\usepackage{courier} 
\usepackage[hyphens]{url} 
\usepackage{graphicx} 
\usepackage{graphicx} 
\usepackage{natbib} 
\usepackage{caption} 
\usepackage{algorithm}
\usepackage{algorithmic}
\usepackage{newfloat}
\usepackage{color}
\usepackage{listings}
\usepackage{amsmath}
\usepackage{stmaryrd}
\usepackage{colonequals}
\usepackage{sansmath}
\usepackage{xcolor}
\usepackage{subcaption}
\usepackage{graphbox}
\usepackage{booktabs}
\DeclareCaptionStyle{ruled}{labelfont=normalfont,labelsep=colon,strut=off} 
\floatstyle{ruled}
\newfloat{listing}{tb}{lst}{}
\floatname{listing}{Listing}

\title{ProbLog4Fairness: A Neurosymbolic Approach to Modeling and Mitigating Bias} 
\author {
    Rik Adriaensen\equalcontrib\textsuperscript{\rm 1},
    Lucas Van Praet\equalcontrib\textsuperscript{\rm 1},
    Jessa Bekker\textsuperscript{\rm 1}, \\
    Robin Manhaeve\textsuperscript{\rm 1},
    Pieter Delobelle\textsuperscript{\rm 1,2},
    Maarten Buyl\textsuperscript{\rm 3}
}
\affiliations {
    \textsuperscript{\rm 1}KU Leuven, Leuven, Belgium\\
    \textsuperscript{\rm 2}Aleph Alpha GmbH, Heidelberg, Germany\\
    \textsuperscript{\rm 3}Ghent University, Ghent, Belgium\\
    \{rik.adriaensen, lucas.vanpraet\}@kuleuven.be
}

\newcommand{\phr}{\beta_R^{hist}} 
\newcommand{\phq}[1]{\beta_{Q,#1}^{hist}} 
\newcommand{\pmr}{\beta_R^{measure}} 
\newcommand{\pmq}[1]{\beta_{Q,#1}^{measure}} 
\newcommand{\pnr}{p_R^{noise}} 
\newcommand{\pnq}[1]{p_{Q,#1}^{noise}} 
\newcommand{\pny}{p^{noise}_Y} 
\newcommand{\pl}{\beta^{label}_Y} 
\newcommand{\dom}[1]{\mathrm{dom}(#1)}

\newtheorem{example}{Example}

\newtheorem{snippet}{ProbLog Template}

\begin{document}

\maketitle

\begin{abstract}
Operationalizing definitions of fairness is difficult in practice, as multiple definitions can be incompatible while each being arguably desirable. Instead, it may be easier to directly describe algorithmic bias through ad-hoc assumptions specific to a particular real-world task, e.g., based on background information on systemic biases in its context. Such assumptions can, in turn, be used to mitigate this bias during training. Yet, a framework for incorporating such assumptions that is simultaneously principled, flexible, and interpretable is currently lacking.
Our approach is to formalize bias assumptions as programs in \textit{ProbLog}, a probabilistic logic programming language that allows for the description of probabilistic causal relationships through logic. Neurosymbolic extensions of ProbLog then allow for easy integration of these assumptions in a neural network's training process. We propose a set of templates to express different types of bias and show the versatility of our approach on synthetic tabular datasets with known biases. Using estimates of the bias distortions present, we also succeed in mitigating algorithmic bias in real-world tabular and image data. We conclude that ProbLog4Fairness outperforms baselines due to its ability to flexibly model the relevant bias assumptions, where other methods typically uphold a fixed bias type or notion of fairness.
\end{abstract}

\section{Introduction}
One way to address fairness concerns is by including a fairness constraint in the learning process~\citep{zafar2017constraints, cruz2024unprocessing}. However, there is no universally accepted definition of fairness, resulting in a large variety of fairness constraints~\cite{verma2018fairness}, some of which are mutually incompatible~\cite{kleinberg2016inherent,chouldechova2017fair,defrance2023maximal}. The choice of fairness constraint can therefore be normatively contentious~\cite{friedler2021possibility}. Moreover, while enforcing or optimizing for the satisfaction of these constraints can be effective in improving the corresponding fairness metric, it lacks the nuance to holistically consider all forms of bias in a specific use case~\cite{selbst2019fairness,buyl2024inherent}.

Alternatively, we can focus directly on addressing the factors that may cause algorithms to be unfair. We assume a significant contributor to such unfairness is the presence of \textit{biasing mechanisms} in the data generation process, as models trained on biased data typically inherit these biases~\cite{mehrabi2021survey}. If we can explicitly model and adjust for these mechanisms during the learning process, the resulting model may `naturally' produce fairer decisions, without ever needing to decide on a specific fairness constraint to optimize.

For example, consider a model that rates the creditworthiness of loan applicants. We may observe that the model scores applicants less favorably in correlation with a certain \textit{sensitive variable} (e.g., ethnicity or gender), which could be considered indirect discrimination~\cite{hacker2018teaching}. This could arise, for instance, because the applicant's place of residence is used in the creditworthiness estimation. However, to what extent is such a correlation justifiable, e.g., because people in the disadvantaged areas are indeed less likely to repay their loans, and how much may this correlation reflect stereotyping, e.g., against an ethnicity dominant in that region? To address these questions, we consider the different mechanisms that may lead to such correlations. If our data consists of labels assigned by human annotators who themselves estimate creditworthiness, they may introduce \textit{label bias} by undervaluing certain groups. We also consider \textit{measurement bias} and \textit{historical bias}, but our method is not limited to these three mechanisms. 

These mechanisms and the data generation process are often considered probabilistic. Thus, it is natural to describe assumptions about them using probabilistic causal models, often represented as Bayesian networks (BNs). However, practitioners report that such models can be difficult to interpret and integrate into data pipelines~\cite{ferrara2024fairness}. Hence, we use ProbLog to declaratively specify the relevant BN in a principled, flexible, and interpretable way using logical facts and rules. DeepProbLog~\cite{manhaeve2021deepproblog}, an extension of ProbLog that allows the parameters of the BN to be predicted by neural networks, then facilitates the integration of these assumptions in the training process of a classifier.
Concretely, this work contributes:
\begin{itemize}
    \item A set of ProbLog templates for describing label, measurement, and historical biasing mechanisms, and an approach for integrating these mechanisms into the training process of a classifier using DeepProbLog. 
    \item Experiments demonstrating the ability of our method to flexibly model the correct bias assumption, thereby significantly improving both fairness and accuracy on unbiased labels, despite training on biased data. 
\end{itemize}

\section{Related Work} 
Research on fairness in machine learning often focuses on quantifying fairness with a metric and then either preprocessing the dataset~\citep{feldman2015certifying,kamiran2012data}, changing the model's optimization process (inprocessing)~\citep{kamishima2011fairness,delobelle2021ethical}, or postprocessing the predictions~\citep{hardt2016equality} to improve the chosen metric.

Some works have used causal models to identify how a sensitive variable affects other features and the final decision~\citep[][\emph{inter alia}]{calders2010three, kilbertus2017, madras2018,kusner2017counterfactual}. For example, \citet{kilbertus2017} model the generation process of biased data as a causal graph and provide a procedure to remove a specific kind of discrimination. \citet{madras2018} model a causal graph containing the decision and its effect, with the sensitive variable as a confounding variable. Based on that graph, they infer the effects of other hidden confounders and learn a fairer decision policy.

Neurosymbolic techniques for fairness build on the idea of modeling and learning causal graphs to obtain fair classifiers. \citet{VARLEY2021106715} introduce an algorithm that learns dependencies between sensitive variables and other variables using a sum-product network and then preprocesses the dataset to remove these dependencies. \citet{Choi2020GroupFB} define a new probabilistic circuit structure that inherently satisfies a set of independence assumptions. \citet{wagner2021} develop a neurosymbolic active learning method based on Logic Tensor Networks and explainability methods, where user queries a trained Logic Tensor Network or tests Shapley values to automatically add constraints to the model that are optimized during retraining.

Recently, \citet{verreet2024pu} have shown that a variety of labeling mechanisms, rules that describe the labeling process, from the Positive-Unlabeled learning setting can be expressed as templates in ProbLog. They show that by expressing their underlying assumptions as ProbLog rules they generalize other methods.
\section{Algorithmic Fairness}
\subsubsection{Notation} Random variables are written as uppercase letters, with the corresponding lowercase letter denoting a specific assignment. Letters in boldface denote vectors. We use the $\Tilde{V}$ notation to indicate a biased version of the variable $V$, where observations of the actual $V$ are typically unavailable.

The binary variable $A$ denotes the sensitive variable (i.e, the variable we want to avoid discrimination against), where $A = 1$ is the sensitive group. The vector $\mathbf{X}$ represents the feature vector, which includes the sensitive variable. Binary variable $Y$ denotes the unbiased label, where $Y = 1$ is favorable. The predicted label is represented by $\hat Y$.

\subsubsection{Defining fairness} Algorithmic fairness is a complex, multi-faceted, and often context-dependent problem with no general solutions~\citep{selbst2019fairness}. Here, we consider the technical challenge of mitigating bias to achieve fair predictions. We introduce some key concepts and refer to \citet{caton2024fairnesssurvey} for further background. 

A common approach to algorithmic fairness is to enforce fairness constraints. For example, \textit{statistical parity}
\begin{equation}
    P(\hat Y = 1 \mid A = 1) = P(\hat Y = 1 \mid A = 0)
\end{equation}
which asserts that both groups have the same positive rate in expectation, or \textit{equalized odds}, which checks whether the true and false positive rates are equal.

Statistical \textit{dis}parity, the difference between the mean positive probability assigned to the sensitive and non-sensitive group, is a measure of fairness derived from this first constraint. 
 
\subsubsection{Fair classifiers} Ignoring fairness considerations, the goal in a binary classification setting is to learn a classifier $h^*: \dom{\mathbf{X}} \to [0, 1]$ predicting the conditional probability $P(\hat Y \mid \mathbf{X})$ that minimizes the prediction error $P(\hat{Y} \neq Y)$. For example, a model trained to minimize binary cross-entropy will approach $h^*$, given sufficient data and capacity. 

However, we consider the scenario in which unbiased features $\mathbf{x}$ and/or labels $y$ are unavailable during training. In practice, datasets are often biased due to, for example, more labeling errors or lower-quality data for certain groups, as well as systemic biases in the past. Therefore, we assume that we observe features $\Tilde{\mathbf{x}}$ and/or labels $\Tilde{y}$ drawn from a biased distribution. Training a classifier to minimize prediction error on this biased data would result in the classifier inheriting these unfair patterns or producing lower-quality predictions for a disadvantaged group.

Therefore, our objective is to learn a function $h$ that minimizes the prediction error as if it were measured on the unbiased dataset, even though we only have biased features $\Tilde{\mathbf{x}}$ and/or labels $\Tilde{y}$ available during training. To achieve this, we assume that the \textit{biasing mechanism}, the mapping between biased and unbiased variables, can be modeled as a BN. 

At test time, it is always the goal to predict unbiased labels. However, we may or may not have access to the unbiased features. Therefore, we explicitly distinguish between two scenarios, where predictions are made on either (1) unbiased features at test time or (2) biased features at test time. 

\section{Deep Probabilistic Logic Programming}
Our method uses \textit{Deep Probabilistic Logic Programming}, an approach to neurosymbolic AI that combines logic programming, probability theory, and neural networks.

\subsection{Logic Programming}
Logic programming is a declarative way of describing a problem in terms of a logical theory, called a logic program, and a query. The task is to determine the truth value of the query within the program. We introduce some fundamental concepts of logic programming here and refer to \citet{flach2022simply} for further background. 

A logic program consists of a set of rules. Rules are expressions of the form $\mathsf{h} \colonminus \mathsf{b_1}, ..., \mathsf{b_k}$, where $\mathsf{h}$ is a logical atom and $\mathsf{b_i}$ are literals, i.e., atoms or their negation. The head $\mathsf{h}$ is true if all $\mathsf{b_i}$'s in the body are true. For example, $\mathsf{wet} \colonminus \mathsf{exposed, raining}$ is a rule expressing that a place is wet if it is exposed and it is raining. A rule with an empty body (e.g., $\mathsf{raining}$) is a fact and automatically true.  In general, atoms are expressions of the form $\mathsf{q(t_1, ..., t_k)}$ potentially containing terms as arguments. These terms can be variables to allow for more general expressions. For example, $\mathsf{wet(P)} \colonminus \mathsf{exposed(P), raining}$ expresses any person $P$ who is exposed will get wet when it is raining. When proving a query, $P$ will be substituted by a specific person.

\subsection{ProbLog}
ProbLog~\citep{deraedt2015probabilistic} is an extension of the logic programming language Prolog with probabilistic facts $\mathsf{p :: f}$, where $\mathsf{f}$ is a fact without variables and $\mathsf{p}$ the probability of that fact being true.
A probabilistic rule $\mathsf{p :: h \colonminus b}$ represents that $\mathsf{h}$ has a probability $\mathsf{p}$ of being true \textit{if} $\mathsf{b}$ is true, thus expressing a conditional probability.

\begin{example}
As an example, consider this ProbLog program modeling a form of label bias.
\[
\begin{aligned}[t]
&\mathsf{poor\_neighborhood(mary).}\\
&\mathsf{can\_pay\_loan(mary).}\\
&\mathsf{can\_pay\_loan(john).}\\
&\mathsf{0.1 :: neg\_bias(A) \colonminus poor\_neighborhood(A).}\\
&\mathsf{gets}\_loan(A) \colonminus can\_pay\_loan(A), \neg neg\_bias(A).\\
\end{aligned}
\]
An applicant $\mathsf{A}$ is given a loan if they can pay it back and there is no negative bias, i.e., a bias causes $10\%$ of all people in a poor neighborhood who can repay a loan to be rejected.
\label{example:problog}
\end{example}

In ProbLog, every probabilistic fact corresponds to an independent Bernoulli random variable that is true with probability $\mathsf{p}$. This induces a probability distribution over all subsets of facts, each of which is called a possible world. The success probability of a query (e.g.,  $\mathsf{gets}\_loan(mary)$) is then defined as the expected probability that a possible world entails that query. In Example \ref{example:problog}, the probability of $\mathsf{gets}\_loan(mary)$ is $0.9$, while the probability of $\mathsf{gets}\_loan(john)$ is $1.0$.

\subsection{DeepProbLog}
DeepProbLog~\citep{manhaeve2021deepproblog} extends ProbLog by allowing the labels of probabilistic facts to be predicted by neural networks, which enables reasoning on top of a neural network's predictions. Example \ref{example:classifier} shows a neural network $\mathsf{h(\mathbf{X})}$ predicting the probability of a label $\mathsf{y_h(\mathbf{X})}$, where $\mathbf{X}$ will be substituted by a specific feature vector $\mathbf{x}$. In Example \ref{example:problog}, for instance, a classifier could predict the probability of $\mathsf{can\_pay\_loan(mary)}$.
\begin{example} A classifier\ as\ a\ probabilistic\ fact.
\[
\begin{aligned}
&\mathsf{h(\mathbf{X})}  \mathsf{:: y_h(\mathbf{X}).}
\end{aligned}
\]
\label{example:classifier}
\end{example}
 In practice, the logic program is compiled into a circuit that computes the probability of a queried fact, e.g., $\mathsf{can\_pay\_loan(mary)}$, given the network's predictions. Importantly, the gradients of distant supervision on entailed facts, e.g., $\mathsf{gets}\_loan(mary)$, can be propagated back through this circuit, which allows the network to be trained with standard gradient-based approaches. 

\section{ProbLog Bias Modeling and Mitigation}

To mitigate algorithmic bias, we now propose a two-step method: (1) model the biasing mechanism as a ProbLog program, and (2) integrate it into the classifier’s training with DeepProbLog's distant supervision to mitigate the bias.

We define a logic program containing both a classifier for predicting the unbiased label $\mathsf{y_h(\mathbf{X})}$ from the unbiased features $\mathsf{\mathbf{X}}$ as in Example \ref{example:classifier} and a biasing mechanism to represent the probabilistic transformation between unbiased features/labels and observed (potentially biased) features/labels. This allows the classifier to learn, from a biased dataset, how to make unbiased classifications, while the mechanism does the transformations between unobserved and observed variables.
The classifier is trained using distant supervision: the program is supervised only through the observed biased labels and features, while DeepProbLog backpropagates gradients through the logic to update the network. This results in gradient updates according to all possible unbiased interpretations consistent with the observed biased data.

If the unbiased features are available at test time, we can drop the logic and use $\mathsf{h(\mathbf{X})}$ directly to make unbiased classifications from the unbiased features. However, if only biased features are available at test time, we keep the mechanism transforming biased to unbiased features. 

DeepProbLog thus offers a ready-to-use framework for mitigating any biasing mechanism modeled in ProbLog including settings with non-tabular data. This strategy, called \textit{ProbLog4Fairness}, is more interpretable than causal fairness methods and allows for flexibly adding or revising assumptions where necessary rather than relying on fully specified graphs. In the remainder of this section we discuss how to model different types of bias.

\subsection{Modeling the Biasing Mechanism}

Although any bias representable as a BN can be modeled in ProbLog~\citep{deraedt2015probabilistic}, we focus on three common bias mechanisms depicted in Figure \ref{fig:BN} to showcase our approach. We assume that the unbiased variables ($X_i$'s and $Y$) are independent of $A$, unless stated otherwise. Since $A$ in that case does not help in predicting $Y$, we generally omit it from the input of a classifier.

\subsubsection{Bias as probabilistic facts}
A biasing mechanism describes the probabilistic transformation between the biased and unbiased distributions of a variable given the sensitive variable. For binary variables, this transformation is fully defined by four probabilities, essentially defining a conditional probability table.
Therefore, we represent bias using four probabilistic facts with probabilities depending on (1) whether the bias affects the value negatively (i.e., changes 1 to 0) or positively (i.e., changes 0 to 1), and (2) whether the person is part of the sensitive group. Example \ref{example:bias-prob-facts} shows these facts for label bias, where variable $\mathbf{X}$ will be substituted for a specific feature vector $\mathbf{x}$. The atom $\mathsf{a(\mathbf{x})}$ is then true when $A = 1$ in that feature vector.

If we assume a bias case never occurs, we can set that probability to 0 or remove the rule. In Example \ref{example:problog}, we only used the first rule with $\mathsf{p_1}=0.1$ to represent the probability of negative discrimination against applicants from a poor neighborhood. Similarly, $\mathsf{p_3}$ in the third rule would represent the probability of someone from a poor neighborhood receiving a loan they cannot repay.

The parameters $\mathsf{p_i}$ can be either set using domain knowledge or estimated from a small subset of the data with both biased and unbiased labels, as demonstrated in the experiments. When the parameters are learned jointly, however, the optimal classifier becomes unidentifiable because the model that minimizes the loss is no longer unique. This is a well-known problem in the PU learning literature \citep{gerych2022propensity_score}. Since these new solutions could contain biased classifiers we consider this setting to be out of scope.

To model bias on a multiclass label or with a categorical sensitive variable we could define a probabilistic fact for every possible transformation between values of the label and a rule for every value of the sensitive variable.

\begin{example} The probabilistic facts for label bias.
\[
\begin{aligned}
&\ \mathsf{p_{1} :: label\_neg\_bias(\mathbf{X}) \colonminus a(\mathbf{X}).}\\
&\ \mathsf{p_{2} :: label\_neg\_bias(\mathbf{X}) \colonminus \neg a(\mathbf{X}).}\\
&\ \mathsf{p_{3} :: label\_pos\_bias(\mathbf{X}) \colonminus a(\mathbf{X}).}\\
&\ \mathsf{p_{4} :: label\_pos\_bias(\mathbf{X}) \colonminus \neg a(\mathbf{X}).}
\end{aligned} 
\]
\label{example:bias-prob-facts}
\end{example}
\subsubsection{Label Bias}
\begin{figure}[t!]
    \centering
    \begin{subfigure}[t!]{0.32\linewidth}
        \includegraphics[width=0.9602\linewidth]{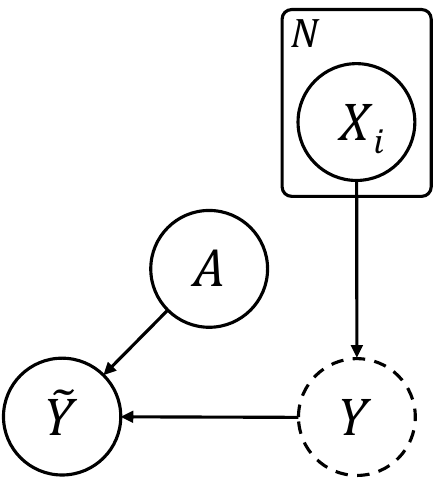}
        \caption{Label}
        \label{fig:BN-label}
    \end{subfigure}
    \begin{subfigure}[t!]{0.32\linewidth}
        \includegraphics[width=\linewidth]{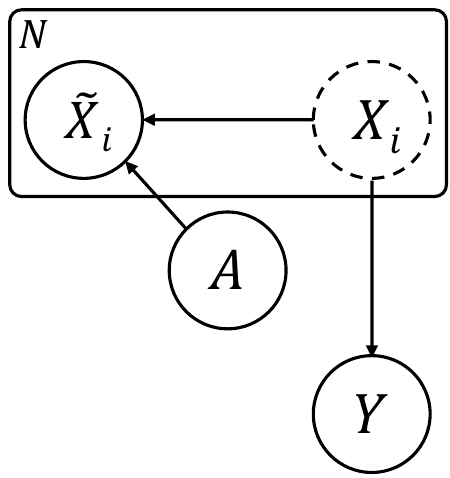}
        \caption{Measurement}
        \label{fig:BN-measure}
    \end{subfigure}
    \hfill
    \begin{subfigure}[t!]{0.32\linewidth}
        \includegraphics[width=\linewidth]{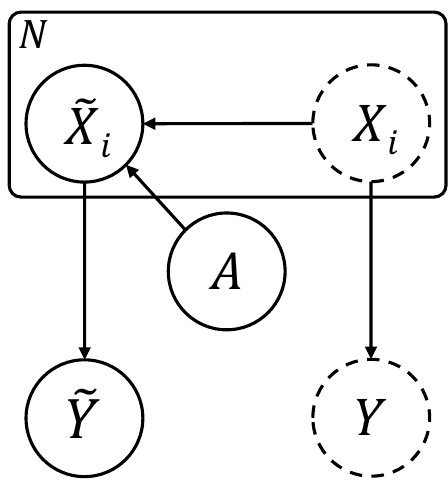}
        \caption{Historical}
        \label{fig:BN-hist}
    \end{subfigure}
    \caption{The Bayesian networks for label, measurement, and historical bias. Dashed nodes are unobserved.}
    \label{fig:BN}
\end{figure}

In the case of label bias (Figure \ref{fig:BN-label}), the unbiased features $\mathbf{X}$ are available and the observed label $\Tilde{Y}$ is a biased and noisy proxy of the unobserved fair label $Y$. In expectation, the labels $\Tilde{Y}$ are less favorable for the sensitive group than the original labels $Y$. This could occur when loan applications are more often rejected because a loan officer directly discriminates against a sensitive variable, such as ethnicity.

The biasing mechanism, i.e., the transformation from $Y$ to $\Tilde{Y}$, is now fully defined by the parameters $P(\Tilde{Y} \mid Y = y, A = a)$ for all $y$ and $a$, as they describe the full conditional probability table of \textit{the biased labels given the unbiased labels}. Template \ref{snip:label-bias} shows the label biasing mechanism in ProbLog, with the classifier and bias probabilities as described in Examples \ref{example:classifier} and \ref{example:bias-prob-facts}. The biased prediction $\mathsf{\Tilde y}$ is true if $\mathsf{y_h}$ is false and the label is positively biased, or if $\mathsf{y_h}$ is true and the label is not negatively biased.

\begin{snippet} Biased prediction under label bias.

\[
\begin{aligned}
&\ \mathsf{\Tilde{y}(\mathbf{X})} \mathsf{~\colonminus \neg y_h(\mathbf{X}) , label\_pos\_bias(\mathbf{X}).}\\
&\ \mathsf{\Tilde{y}(\mathbf{X})} \mathsf{~\colonminus y_h(\mathbf{X}) , \neg label\_neg\_bias(\mathbf{X}).}
\end{aligned}
\]
\label{snip:label-bias}
\end{snippet}

 A concrete program can model multiple mechanisms that affect the label or it can make stricter assumptions, for example, by assuming the biased labels are always correct for the non-sensitive group. This lowers the number of parameters needed to describe the full conditional probability table and simplifies the mechanism, as it is equivalent to removing lines 2 and 4 in Example \ref{example:bias-prob-facts}. For instance, Example \ref{example:label-bias} shows a loan application case in which the only source of bias is a negative bias for the sensitive group, and there is noise on the label due to random mistakes with probability $\mathsf{p_{noise}}$. 

\begin{example} A specific model for label bias in a loan application.
\[
\begin{aligned}
&\ \mathsf{h(\mathbf{X})}  \mathsf{\ :: can\_pay\_loan(\mathbf{X}).}\\
&\ \mathsf{0.21 :: neg\_bias(\mathbf{X}) \colonminus poor\_neighborhood(\mathbf{X}).}\\
&\ \mathsf{gets}\_loan(\mathbf{X}) \mathsf{~\colonminus can\_pay\_loan(\mathbf{X})}, \mathsf{\neg neg\_bias(\mathbf{X}).}\\
&\ \mathsf{p_{noise} :: noise(\mathbf{X})}.\\
&\ \mathsf{observed\_gets}\_loan(\mathbf{X}) \mathsf{~\colonminus gets}\_loan(\mathbf{X}), \mathsf{\neg noise(\mathbf{X})}.\\
&\ \mathsf{observed\_gets}\_loan(\mathbf{X}) \mathsf{~\colonminus \neg gets}\_loan(\mathbf{X}), \mathsf{noise(\mathbf{X})}.
\end{aligned}
\]
\label{example:label-bias}
\end{example}

\subsubsection{Measurement Bias} When there is measurement bias (Figure \ref{fig:BN-measure}), the observed features are noisy and biased proxies of the unbiased features, but the label only depends on the unbiased features. Unfairness arises when these proxy features are often less favorable or noisier for the sensitive group. For example, measurement bias occurs when `days worked in the past three years' is used as a proxy for job stability, which is negatively biased against women who took maternity leave.

The parameters describing the bias are now specified by the conditional probability table of some biased feature $\Tilde{X}_i$ given the unbiased feature $X_i$, i.e., $P(X_i \mid \Tilde{X}_i = \Tilde{x}_i, A = a)$ for all $\Tilde{x}_i$ and $a$. The probabilistic facts follow the same structure as in Example \ref{example:bias-prob-facts} but describe the transformation \textit{from a biased to an} unbiased feature, opposite to before. The mechanism for predicting the unbiased label under measurement bias is given in Template \ref{snip:measure-bias} for a feature vector with biased feature $\mathsf{b}$. 

\begin{snippet} Fair prediction under measurement bias.
\[
\begin{aligned}
&\ \mathsf{b\_biased(\Tilde{\mathbf{X}})} \mathsf{~\colonminus \neg b(\Tilde{\mathbf{X}}) , b\_neg\_bias(\Tilde{\mathbf{X}}).}\\
&\ \mathsf{b\_biased(\Tilde{\mathbf{X}})} \mathsf{~\colonminus b(\Tilde{\mathbf{X}}) , b\_pos\_bias(\Tilde{\mathbf{X}}).}\\
&\ \mathsf{debias(\Tilde{\mathbf{X}}, \mathbf{X})} \mathsf{~\colonminus b\_biased(\Tilde{\mathbf{X}}), debias\_b(\Tilde{\mathbf{X}}, \mathbf{X}).}\\
&\ \mathsf{debias(\Tilde{\mathbf{X}}, \Tilde{\mathbf{X}})} \mathsf{~\colonminus \neg b\_biased(\Tilde{\mathbf{X}}).}\\
&\ \mathsf{y(\Tilde{\mathbf{X}})}  \mathsf{~\colonminus debias(\Tilde{\mathbf{X}}, \mathbf{X}), y_h(\mathbf{X}).}\\
\end{aligned}
\]
\label{snip:measure-bias}
\end{snippet}

\begin{figure*}[ht!]
    \centering
    \includegraphics[width=0.94\textwidth]{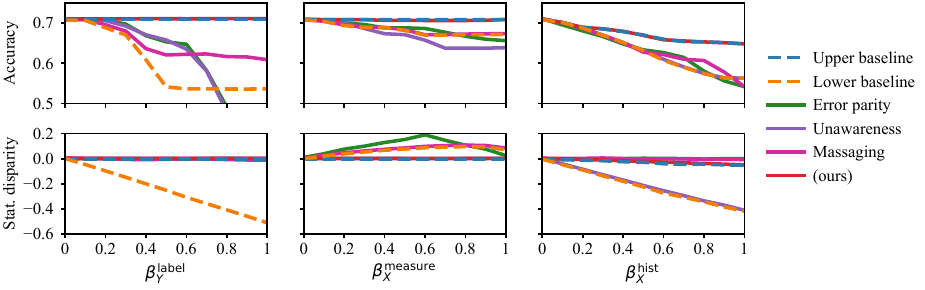}
    \caption{\textbf{ProbLog4Fairness successfully models different types of bias, approaching the upper baseline.} We measure accuracy and statistical disparity for an increasing probability of label (left), measurement (middle), and historical (right) bias during training while evaluating on unbiased data at test time.}
    \label{fig:bod-bias-sweep}
\end{figure*}

The label prediction is the combination of the predictions on the different possible unbiased feature vectors, weighted by how probable they are given the biased feature vector. Predicate $\mathsf{b(\Tilde{\mathbf{X}})}$ is true when the biased feature in the feature vector $\mathbf{\Tilde{X}}$ is one and $\mathsf{debias\_b(\Tilde{\mathbf{X}}, \mathbf{X})}$ substitutes variable $\mathbf{X}$ with a vector identical to $\Tilde{\mathbf{X}}$ but with the value of $\mathsf{b}$ flipped. This approach to modeling bias on $\mathsf{b}$ can be easily extended to model multiple biased features.

\subsubsection{Historical Bias} 
Historical bias (Figure \ref{fig:BN-hist}) occurs when the features in the dataset are biased and, in contrast to measurement bias, the labels are based on the biased features. Therefore, the bias is present in both the features and the label.
For example, discrimination can lead to limited job opportunities, resulting in less favorable features (e.g., lower income or unemployment), which actually cause the sensitive group to default on loans more often. However, in a normatively desirable world without discrimination, loan repayment would be more consistent as these features would be more similarly distributed for the sensitive group. The transformation from the historically biased to the unbiased distribution can be modeled in the same way as measurement bias in Template \ref{snip:measure-bias}. As the labels would also change in this desired world, different from measurement bias, we can combine this with a transformation from unbiased to biased labels modeled as label bias in Template \ref{snip:label-bias}.

However, it is sometimes possible under historical bias to assume the function from biased features to biased labels is the same as the one from unbiased features to unbiased labels. As a result, a classifier trained on the biased features and labels will make fair predictions if the features are unbiased at test time. Under this assumption, the bias is \textit{only} a result of the unfavorable feature distribution for the sensitive group. Therefore, if we have biased features at test time, we can simply learn the fair classifier on the biased data and use the measurement biasing mechanism during test time to predict as if in a normatively desirable world. This can help counteract the historical disadvantages, for example, by providing people who were historically discriminated against with more opportunities despite having a lower income. 

\section{Experiments}
We now verify our bias modeling and mitigation approach \textit{ProbLog4Fairness} experimentally against synthetic data and two real-world datasets: one tabular and one image.

\subsubsection{Baselines.}
All baselines use a neural network as a classifier with an architecture as described in Appendix~\ref{app:experimental-details}. The \textit{Lower baseline} is directly trained on the observed features and labels. The \textit{Upper baseline} is directly trained on the unbiased features and labels. \textit{Unawareness} is trained on the observed features and labels, but with omission of the sensitive variable~\citep{dwork2012fairness}. \textit{Massaging} preprocesses the biased dataset by demoting positive labels from the non-sensitive group and promoting negative labels from the sensitive group based on the ranking of a model fitted to the data. Afterwards, a classifier is trained on this massaged dataset~\citep{kamiran2009}. Finally, \textit{Error parity} is a postprocessing technique that adapts the lower baseline's predictions to satisfy a fairness constraint~\citep{cruz2024unprocessing}. We use statistical parity as fairness constraint for this baseline.

\subsubsection{Methodology.} To measure predictive performance, we report accuracy or the F1 score when labels are imbalanced. We use statistical disparity to evaluate fairness. We tested using 5-fold cross-validation and repeated each experiment on at least five different seeds, reporting the mean. Further experimental details and hyperparameters are discussed in Appendix \ref{app:experimental-details}.

\subsection{Synthetic Data Experiments}
First, we experiment on synthetic datasets where we have explicit control over the type and probability of bias. The data generation process we use is based on~\citep{baumann2023biasondemand}, but restricted to binary and categorical features. The probability of label, measurement, and historical bias occurring is controlled by $\beta^{label}_Y$, $\beta^{measure}_\mathbf{X}$, and $\beta^{hist}_\mathbf{X}$, respectively. 
If there is measurement or historical bias, the probability is assumed to be the same for each feature. Further details on data generation can be found in Appendix \ref{app:synth-data-gen}. 
For RQ1 and RQ2, we assume the relevant bias probability $\beta$ is known and we set the parameters in our program accordingly. In RQ3, we investigate what to do when $\beta$ is unknown.

\paragraph{RQ1.} \textbf{How does ProbLog4Fairness compare to the baselines under different types of bias} in terms of predictive performance and fairness? Figure \ref{fig:bod-bias-sweep} shows the accuracy and statistical disparity of our approach against the baselines under label bias (using Template \ref{snip:label-bias} during training), measurement bias (using Template \ref{snip:measure-bias} during training) and historical bias (using Template \ref{snip:measure-bias} at test time) for an increasing bias probability. The models are tested on unbiased data. Given the correct parameters for the program, our approach achieves an accuracy and statistical disparity comparable to the upper baseline despite training on the biased labels. The mitigating baselines are generally effective in achieving fairness, but their predictive performance on the unbiased data is limited. Due to the flexibility of our programs, we are able to deal with a wide range of biases effectively.
\begin{figure}[t!]
    \centering
    \includegraphics[width=\linewidth]{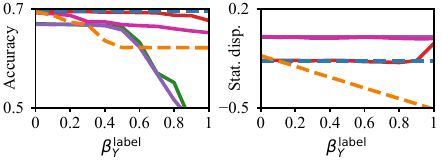}
    \caption{\textbf{Our method is able to remove only the problematic bias when $A \not\perp Y$, approaching the upper baseline.} We measure accuracy and statistical disparity on unbiased data while training with an increasing probability of label bias under $A \not\perp Y$. Color coding is consistent with the legend in Figure \ref{fig:bod-bias-sweep}}
    \label{fig:bod-bias-sweep-dep}
\end{figure}
\paragraph{RQ2.} \textbf{Can our approach mitigate bias when $A \not\perp Y$?} ProbLog4Fairness is able to effectively mitigate only the problematic bias, as can be seen in Figure \ref{fig:bod-bias-sweep-dep}. By setting the parameters of our program based on the correct $\beta$, which does not capture the nonproblematic part of the correlation between $A$ and $Y$, we achieve the statistical disparity of the upper baseline, which is no longer expected to be close to zero. In contrast, the lower baseline predicts a disproportionately smaller mean positive probability for the sensitive group as the probability of the bias increases. The other baselines are not able to make the distinction between problematic and nonproblematic bias and incorrectly impose a statistical disparity of zero.

\begin{figure}[t!]
    \center{
    \includegraphics[width=\linewidth]{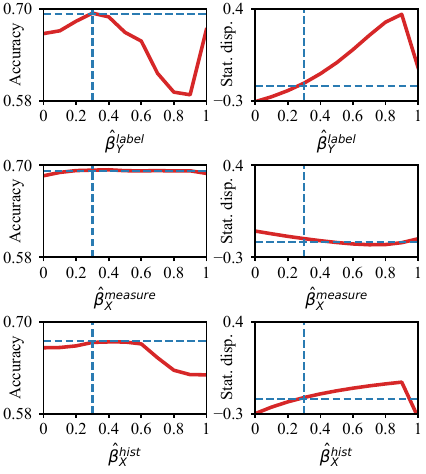}
    }
    \caption{
    \textbf{Our method achieves the highest accuracy and the expected statistical disparity when the correct bias probability is used to estimate the parameters in the program.} We train on fixed bias probabilities of 0.3, with $A \not \perp Y$ and evaluate on unbiased data, but vary the bias probability $\hat \beta$ used to set the program's parameters. The dashed lines indicate the upper baseline.}
    \label{fig:bod-assumption-sweep}
\end{figure}
\paragraph{RQ3.} \textbf{What is the effect of choosing the wrong parameters in the program?}
Figure \ref{fig:bod-assumption-sweep} shows the statistical disparity and accuracy for our method, evaluated on unbiased labels, for a fixed label bias probability of 0.3 and $A \not \perp Y$, as we vary the bias probability $\hat \beta$ used to calculate the parameters of the program. The trained classifier achieves the highest accuracy and a statistical disparity closest to the upper baseline when the parameters in the program are chosen to match the actual bias probability. Importantly, around this optimum, the sensitivity for a parameter estimation error is small. Therefore, our method will be effective if we estimate these parameters from a limited subset of the data for which the unbiased features/labels are also available, as discussed in Appendix \ref{app:sample-complexity}. 

\subsection{Student Dataset Experiment} The \textit{Student Alcohol Consumption} dataset introduced by \citet{cortez2008alcoholconsumption} contains tabular features about 856 students, such as their weekly alcohol consumption, gender, and how long they studied for an exam. The labels indicate whether they passed the exam and can be considered the unbiased labels (keeping any historical biases in exam success rates out of scope). Based on this data \citet{lenders2023student} asked annotators to predict whether these students would pass based on their features. By priming the annotators, this produced biased labels against male students. This dataset is useful as the biased and unbiased labels are known, yet the bias is \textit{real-world} in the sense that it results from the actual biased decisions of the human annotators.

\begin{figure}[t!]
    \centering
    \includegraphics[width=\linewidth]{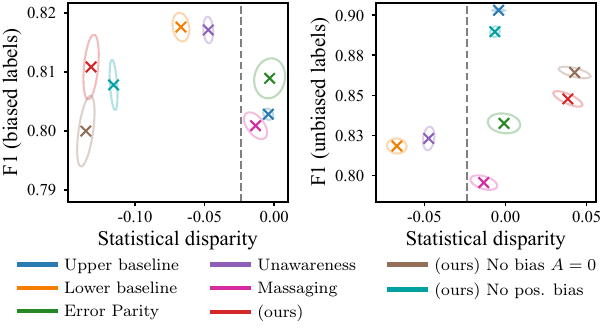}
    \caption{\textbf{Our approach achieves a higher F1 score on the \textit{unbiased} labels than the mitigating baselines and approaches the expected statistical disparity. Sensible simplifying assumptions seem to benefit performance.} The gray vertical line indicates the statistical disparity in the unbiased labels. The ellipses show a 95\% confidence region based on the standard error.}
    \label{fig:student}
\end{figure} 

\paragraph{RQ4.}\textbf{Can ProbLog4Fairness effectively mitigate real-world bias?}
We model the bias as label bias, with $A \perp Y$, and assume the bias probability is known. Figure \ref{fig:student} shows that our method approaches the statistical disparity present in the unbiased data and outperforms the baselines in terms of unbiased label F1 score. As shown in Appendix \ref{app:experimental-results}, the results are similar when equalized odds is used as the fairness metric. When evaluating on biased data, the baselines incorrectly appear to perform better emphasizing the importance of evaluating on unbiased labels. As expected, we do not perform as close to the upper baseline for real-world data, likely because other biases are present that are not being accounted for. Therefore, we expect a more elaborate ProbLog program to achieve an even better F1 score.

\paragraph{RQ5.} \textbf{What is the effect of simplifying assumptions in the program on the predictive performance and fairness of our method?} In Figure \ref{fig:student}, we also show how two ProbLog programs with simplifying assumptions, `no positive bias{'} and `no bias on $A = 0$', perform. The simplified programs achieve a better F1 score and `no positive bias' achieves a better statistical disparity. This could be due to a large estimation error on the positive bias parameters, as there are only 67 girls and 82 boys who did not pass and thus have an unbiased negative label. Another hypothesis favoring simple mitigation strategies is that under an expressive prior, a classifier could have difficulty learning what remains.

\subsection{CELEB-A Experiment}
The CELEB-A dataset~\citep{liu2015celeba} contains faces of celebrities with 40 binary attributes for each image, such as Smiling, Mouth Slightly Open, Blurry, and High Cheekbones. The aim is to predict these features from the image. \citet{wu2023consistencyceleba} show that these labels are often subjective and inconsistent between annotators, resulting in biased labels. They also provide a cleaned version of the Mouth Slightly Open attribute. We fine-tuned a ResNet-50 pretrained on ImageNet~\citep{imagenet} on the cleaned labels as an upper baseline and achieved a 10.18 \% improvement in F1 score on the cleaned labels compared to fine-tuning on the original labels. Although fairness is less of a concern for this particular attribute, bias in facial recognition often does lead to unfair decisions when the label quality for a particular group is lower. 
\begin{figure}[t!]
    \centering
    \includegraphics[width=\linewidth]{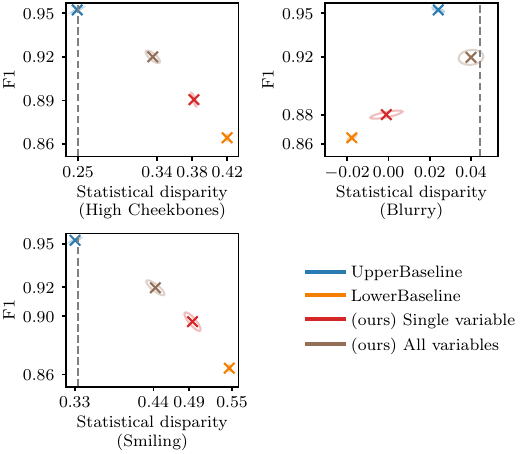}
    \caption{\textbf{Our method successfully mitigates label bias for facial feature detection. Combining biasing mechanisms over multiple attributes benefits the overall F1 score on unbiased Mouth Slightly Open labels.} The dashed lines indicate the actual statistical disparities present in the unbiased data. The ellipses show a 95\% confidence region based on the standard error.}
    \label{fig:celeba}
\end{figure}
\paragraph{RQ6.} \textbf{Can ProbLog4Fairness effectively mitigate label bias on high-dimensional features, such as images?} 
As shown in Figure \ref{fig:celeba}, our method achieves part of the increase in F1 score of the upper baseline under the assumption of label bias on Mouth Slightly Open compared to three `sensitive' variables, Smiling, Blurry, and High Cheekbones that turn out to be correlated with the labeling errors. We also effectively progress the statistical disparity towards the level present in the unbiased data with respect to these three attributes. Notice, however, that correcting for all three variables at the same time does much better in terms of both F1 score and statistical disparity, compared to taking a single variable into account. This shows that eliminating the bias with respect to a single feature does not immediately imply fairness because other features correlated with the sensitive variable are not accounted for. This underlines the importance of being able to flexibly model the large number and variety of biases that can be present in a real-world context. Appendix \ref{app:experimental-results} shows that these results are similar when equalized odds is used as the fairness metric.

\section{Conclusion}
In conclusion, ProbLog4Fairness successfully models and mitigates various types of bias across synthetic and real-world datasets. Due to our ability to flexibly model the relevant bias assumptions, we outperform baselines that uphold a fixed assumption of bias or notion of fairness. Additionally, we show that the parameters in our program, if not available from background information, can be set using a small unbiased subset of the data. 

\section{Acknowledgements}
 This work was partially supported by the Internal Funds KU Leuven iBOF/21/075. This research also received funding from the Flemish Government (AI Research Program), the European Research Council (ERC) under the European Union’s Horizon 2020 research and innovation program (Advanced Grant DeepLog No. 101142702). We are greatful to Luc De Raedt and Tijl De Bie for providing helpful feedback.

\bibliography{citations}

\setcounter{secnumdepth}{1}

\clearpage
\appendix
\section{Experimental Details} \label{app:experimental-details}
\input{app_experimental_details}

\section{Synthetic Data Generation} \label{app:synth-data-gen}
\input{app_synthetic_data}

\section{Sample Complexity Argument} \label{app:sample-complexity}
\input{app_sample_complexity}

\section{Additional Experimental Results} \label{app:experimental-results}
\input{app_additional_experiments}

\end{document}

%% file: app_experimental_details.tex
\begin{table*}[t!]
    \centering
    \begin{tabular}{l|cccccccc}
        \toprule
        \multicolumn{1}{c}{\textbf{Name}} & \textbf{\# Layers} & \textbf{Layer width} & \textbf{Dropout} & \textbf{Lr} &\textbf{ Batch 
        size} & \textbf{Loss} & $\mathbf{A \perp Y}$\textbf{?} & \textbf{\# Seeds}  \\
        \midrule
        Figure \ref{fig:bod-bias-sweep} & 3 & 32 & 0.0 & $3 \times 10^{-4}$ & 64 & BCE Loss & Yes & 5 \\
        Figure \ref{fig:bod-bias-sweep-dep} & 3 & 32 & 0.0 & $3 \times 10^{-4}$ & 64 & BCE Loss & No & 5 \\
        Figure \ref{fig:bod-assumption-sweep} & 3 & 32 & 0.0 & $3 \times 10^{-4}$ & 64 & BCE Loss & No & 5 \\
        Figure \ref{fig:student} & 3 & 256 & 0.2 & $3 \times 10^{-4}$ & 64 & Focal Loss \cite{lin2017focal} & Yes & 10 \\
        Figure \ref{fig:celeba} & \multicolumn{3}{c}{ResNet-50} & $1 \times 10^{-4}$ & 16 & BCE Loss & No & 5 \\
        \bottomrule
    \end{tabular}
    \caption{The neural network architectures and hyperparameters used for each of the figures shown in the main text.}
    \label{tab:exp-details}
\end{table*}

\begin{table*}[t!]
    \centering
    \begin{tabular}{l|cccc}
        \toprule
        \multicolumn{1}{c}{\textbf{Name}} & \textbf{CPU} & \textbf{Memory} & \textbf{GPU}  & \textbf{Reproduction time} \\
        \midrule
        Figure \ref{fig:bod-bias-sweep} &  Intel Xeon Silver 4514Y & 256 Gb & \multicolumn{1}{c}{-} & \phantom{0}$\sim$72h \\
        Figure \ref{fig:bod-bias-sweep-dep} &  Intel Xeon Silver 4514Y & 256 Gb & \multicolumn{1}{c}{-} & \phantom{0}$\sim$22h \\
        Figure \ref{fig:bod-assumption-sweep} &  Intel Xeon Silver 4214 & 256 Gb & \multicolumn{1}{c}{-} & \phantom{0}$\sim$44h  \\
        Figure \ref{fig:student} & Intel Xeon Silver 4214  & 128 Gb & \multicolumn{1}{c}{-} & \phantom{00}$\sim$3h  \\
        Figure \ref{fig:celeba} &  Intel Xeon CPU E5-2690 v3 & 32 Gb & NVIDIA Titan RTX & $\sim$100h \\
        \bottomrule
    \end{tabular}
    \caption{The estimated \textit{sequential} runtime and hardware used for each of the figures shown in the main text.}
    \label{tab:hardware}
\end{table*}

Here we provide full experimental details for each figure shown in the main text.

\paragraph{Evaluation} To measure predictive performance, we report accuracy or F1 score when there is a label imbalance. We use statistical disparity to evaluate fairness. We test using 5-fold cross-validation and repeat each experiment on at least 5 different seeds (see Table \ref{tab:exp-details}), reporting the mean.

\paragraph{Architecture and hyperparameters} For all synthetic data experiments and the student dataset experiment, we use a standard feedforward neural network with ReLU activation functions and dropout, implemented in PyTorch \citep{paszke2017pytorch}, as a classifier. The number of layers and layer width are specified in Table \ref{tab:exp-details}. When the sensitive variable is assumed to be independent of the unbiased label, we omit sensitive features from the input of classifiers trained with ProbLog4Fairness, otherwise, these features are retained. 

For the CELEB-A experiment, we use the ResNet-50 architecture \cite{he2016resnet} with weights pretrained on ImageNet1k-v2 \cite{imagenet}, available through PyTorch. In this case, we substitute the final feedforward layer and keep the whole network trainable, using a low learning rate. We optimize using AdamW \cite{loshchilov2017adamW}, with the learning rates, batch sizes, and loss functions specified in Table \ref{tab:exp-details}. We select the best model based on minimal loss on the standard validation set in the case of CELEB-A, or a random 10 percent validation set held-out from the training data otherwise.

\paragraph{Hardware} The hardware used and the estimated total runtime required to reproduce each figure in the main text are listed in Table \ref{tab:hardware}. The estimated runtimes include all baselines and are computed as the total sequential runtime, even though the experiments were ran in parallel.

\paragraph{ProbLog programs} The full DeepProbLog programs used to train our method 
under label and measurement bias are shown in Listing \ref{lst:label} and \ref{lst:measure}, respectively. For historical bias, we use the measurement bias program at test-time. The program used to mitigate label bias with respect to three sensitive attributes simultaneously in the CELEB-A experiment is shown in Listing \ref{lst:celeba}. 

\begin{listing}[H]
\begin{lstlisting}[frame=topline, caption={Full DeepProbLog programs for mitigating label bias with respect to one sensitive attribute.}, label={lst:label}]
nn(h,X) :: y_h(X).
nn(a,X) :: a(X).

p1 :: label_neg_bias(X) :- a(X).
p2 :: label_neg_bias(X) :- \+a(X).
p3 :: label_pos_bias(X) :- a(X).
p4 :: label_pos_bias(X) :- \+a(X).

y_(X) :- y_h(X), \+label_neg_bias(X).
y_(X) :- \+y_h(X), label_pos_bias(X).

?- y_(x).
\end{lstlisting}
\end{listing}

In the main text, we write $h(\mathbf{X})$ to represent $nn(h, X)$ and $\neg$ to mean $\backslash +$ (ProbLog's negation as a failure symbol). Both can be considered syntactic sugar. The program in Listing \ref{lst:measure} models measurement bias on all $n$ features in the feature vector, different from Snippet \ref{snip:measure-bias}, which only models bias on a single feature $b$. Now, the predicate \lstinline{debias(X_,X,N)} is recursively defined to traverse the feature vector \lstinline{X_}, using \lstinline{N} as an index, to debias all $n$ features. The term \lstinline{x_(0,0,0,0)} is the identifier of the original feature vector, while the term \lstinline{x_(1,0,0,0)}, for example, carries this same feature vector with the value of the first feature flipped.

The \textit{neural predicates} that select attributes from the feature vectors (e.g., $nn(a,X)$) can be seen as non-trainable functions from feature vectors to probabilities and are implemented as a PyTorch module in practice. When we state that ``we consider the parameters of the program known", this means that we either derive their value exactly from the data generation equations, in the case of synthetic data or estimate them from fair data available for the student dataset and CELEB-A experiments. Note that for the experiments under historical bias (Figure \ref{fig:bod-bias-sweep} and \ref{fig:bod-assumption-sweep}), we predict the unbiased labels from the \textit{biased} features at test-time. For other bias types, we always evaluate how well we predict unbiased labels from \textit{unbiased} features.

\begin{listing}[H]
\begin{lstlisting}[frame=topline,caption={Full DeepProbLog programs for mitigating measurement bias with respect to one sensitive attribute.}, label={lst:measure}]
nn(h,X)    :: y_h(X).
nn(n,X_,N) :: n(X_,N).
nn(a,X_)   :: a(X_).

debias_n(1, x_(0,X,Y,Z), x_(1,X,Y,Z)).
debias_n(2, x_(R,0,Y,Z), x_(R,1,Y,Z)).
debias_n(3, x_(R,X,0,Z), x_(R,X,1,Z)).
debias_n(4, x_(R,X,Y,0), x_(R,X,Y,1)).

p1(N) :: n_neg_bias(X_,N):- a(X_).
p2(N) :: n_neg_bias(X_,N):- \+ a(X_).
p3(N) :: n_pos_bias(X_,N):- a(X_).
p4(N) :: n_pos_bias(X_,N):- \+ a(X_).

n_biased(X_,N)                            :-\+n(X_,N),n_neg_bias(X_,N).
n_biased(X_,N)                              :-n(X_,N),n_pos_bias(X_,N).

debias(X_, X_, 0).
debias(X_, X, N):- >(N,0),is(N2,N-1),n_biased(X_,N),debias_n(N,X_,Xf),  debias(Xf,X,N2).
debias(X_, X, N):- >(N,0),is(N2,N-1),       \+n_biased(X_,N),debias(X_,X,N2).

y(X_):- debias(X_,X,4), y_h(X).

?- y(x_(0,0,0,0)).
\end{lstlisting}
\end{listing}

\begin{listing}[H]
\begin{lstlisting}[frame=topline,caption={Full DeepProbLog program to mitigate label bias with respect to the three sensitive attributes as used in the CELEB-A experiment (Figure \ref{fig:celeba}).}, label={lst:celeba}]
nn(h,X) :: y_h(X).

nn(hc,X) :: hc(X).
nn(bl,X) :: bl(X).
nn(sm,X) :: sm(X).

p1_hc :: label_neg_bias_hc(X):- hc(X).
p2_hc :: label_neg_bias_hc(X):- \+hc(X).
p3_hc :: label_pos_bias_hc(X):- hc(X).
p4_hc :: label_pos_bias_hc(X):- \+hc(X).

p1_bl :: label_neg_bias_bl(X):- bl(X).
p2_bl :: label_neg_bias_bl(X):- \+bl(X).
p3_bl :: label_pos_bias_bl(X):- bl(X).
p4_bl :: label_pos_bias_bl(X):- \+bl(X).

p1_sm :: label_neg_bias_sm(X):- sm(X).
p2_sm :: label_neg_bias_sm(X):- \+sm(X).
p3_sm :: label_pos_bias_sm(X):- sm(X).
p4_sm :: label_pos_bias_sm(X):- \+sm(X).

y1(X) :- y_h(X), \+label_neg_bias_hc(X).
y1(X) :- \+y_h(X), label_pos_bias_hc(X).

y2(X) :- y1(X), \+label_neg_bias_bl(X).
y2(X) :- \+y1(X), label_pos_bias_bl(X).

y_(X) :- y2(X), \+label_neg_bias_sm(X).
y_(X) :- \+y2(X), label_pos_bias_sm(X).

?- y_(x).
\end{lstlisting}
\end{listing}

%% file: app_synthetic_data.tex
The data generation process we use is based on \citet{baumann2023biasondemand}, but is restricted to binary features drawn from a Bernouilli distribution B. This is easily extended to categoricals. As in the original paper, we differentiate the features previously denoted as $\mathbf{X}$ into a resource feature (e.g., income) $R$ and $n_Q$ features correlated with the resource (e.g., education) with $Q_i$ for $i \in [1, n_Q]$. However, whereas \citet{baumann2023biasondemand} model bias as an additive effect on an original continuous value, we model it as a probabilistic flipping of the original binary or categorical value. The probability of label, measurement, and historical bias occurring is controlled by $\beta^{label}_Y$, $\beta^{measure}_X$, and $\beta^{hist}_X$, respectively. The probability of measurement and historical bias is assumed to be equal for all features. In addition to these $\beta$ probabilities for introducing \textit{negative bias}, the parameters $p^{noise}_Y$, $p^{noise}_R$ and $p^{noise}_{Q,i}$ control additional noise on the sensitive group.

Below, $\oplus$ denotes the XOR operation and $\llbracket\ \rrbracket$ denotes Iverson brackets, which transform True into 1 and False into 0.

\begin{align}
    &A \sim B(p_A) \\
    &R \sim B(p_R) \land (1- B(\phr)) \\
    &Q_i \sim B(p_{Q,i} + \alpha_{QR, i}R) \land (1- B(\phq{i}A)) \\
    &S = \alpha_A (1 - A) + \alpha_R R + \sum_{i}^{n_Q} \alpha_{Q,i} Q_{i} + N(0, \sigma_y^2) \\
    &Y = \llbracket S > \bar{S} \rrbracket \\
    &\Tilde{R} \sim R \land (1-B(\pmr A)) \oplus B(\pnr) \\
    &\Tilde{Q}_i \sim Q \land (1-B(\pmq{i} A)) \oplus B(\pnq{i}) \\
    &\Tilde{Y} \sim Y \land (1 - B(\pl * A)) \oplus B(\pny)
    \label{eq:syth-data-generation}
\end{align}\\

The unbiased label is 1 if the latent feature $S$ is above the threshold $\bar{S}$, where $S$ is the weighted sum of all features and some unbiased normally distributed noise $N$ with mean 0 and variance $\sigma_y^2$. As in the original paper, the unbiased function the classifier must learn is \textit{linear} from noisy data.

All experiments that use synthetic data generate datasets of size 10{,}000 and share the following parameters: $n_Q = 3$, $p_A = p_R = p_{Q,i} = 0.5$, $\alpha_R = \alpha_{Q,i} = 1$, $\alpha_{RQ,i} = i/10$ and $\sigma_y = 2$. 

In experiments where we introduce label bias (including when that is with probability 0.0), we set $p^{noise}_Y = 0.1$. Equivalently, in experiments where we introduce measurement bias, $p^{noise}_R = p^{noise}_{Q,i} = 0.1$. When we say $A \not\perp Y$, then $\alpha_A = 1$, otherwise, $\alpha_A = 0$. We set $\bar S = 1.5$ when $\alpha_A = 0$ and choose $\bar S = 2.5$ otherwise, to ensure balanced classes.

\citet{baumann2023biasondemand} classify bias along two axes: measurement-historical and feature-label bias. However, in their mathematical model measurement and historical bias on the label are equivalent but represented by a separate parameter to show the different sources the bias can come from (e.g., poorer quality labels for the sensitive group or a systemic bias in the past). We choose to represent all label bias using a single parameter, without loss of generality. Therefore, what the original paper describes as measurement and historical bias on the features, we simply refer to as measurement and historical bias, while our label bias includes both their measurement and historical bias on the label. Note that this grouping of types of label bias is only applied to data generation, it is still entirely possible to model multiple causes of label bias separately in a single ProbLog program.

%% file: app_sample_complexity.tex
Here we discuss how to set the parameters in our programs. Parameters are either set using domain knowledge or estimated from data. In some real-world task, there are studies that quantify the probability and nature of bias introduced due to, e.g., socio-economic factors or limitations in measuring methods. When such prior knowledge is available it can be used to set the parameter values. However, if this is not the case, we might want to estimate the parameters from a small subset of the data for which we have both biased and unbiased version available. 

As shown in Figure \ref{fig:bod-assumption-sweep}, our method is not highly sensitive to parameter estimation errors around the correct value. Therefore, ensuring a small estimation error leads to only a minor decreases in predictive performance and fairness.

To calculate the necessary dataset size for a small estimation error we use a simple sample complexity argument based on Hoeffding's inequality \cite{hoeffding1963inequality}. It tells us that we need $n \geq \frac{1}{2 \varepsilon^2} \ln\left( \frac{2}{1-\gamma} \right)$ samples if we want to estimate a parameter within an error $\epsilon$ with confidence $\gamma$. For example, to estimate the probability of negative label bias for the sensitive group within an error of 10\% with 95\% confidence, we need 184 datapoints in the sensitive group with an unbiased positive label.

%% file: app_additional_experiments.tex
To demonstrate that our results on real-world data are consistent across fairness metrics, we present the observed equalized odds metric for the Student and CELEB-A datasets in the figures below.

\begin{figure}[H]
    \centering
    \includegraphics[width=\linewidth]{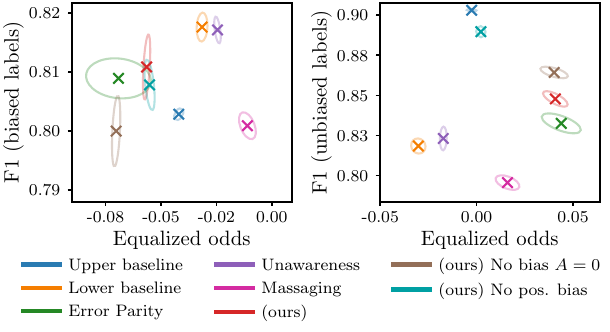}
    \caption{The F1 score from the experiment on the Student dataset (Figure \ref{fig:student}) plotted against equalized odds, showing that the results are consistent across fairness metrics. The ellipses indicate a 95\% confidence region based on the standard error.}
\end{figure} 

\begin{figure}[H]
    \centering
    \includegraphics[width=\linewidth]{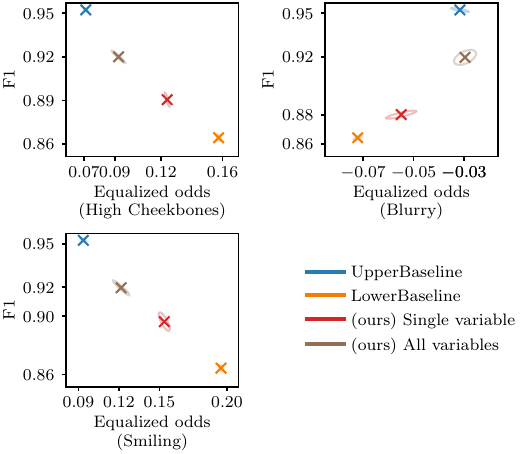}
    \caption{The F1 score from the experiment on the CELEB-A dataset (Figure \ref{fig:celeba}) plotted against equalized odds, showing that the results are consistent across fairness metrics. The ellipses indicate a 95\% confidence region based on the standard error.}
\end{figure}